\documentclass[conference]{IEEEtran}
\IEEEoverridecommandlockouts
\usepackage{cite}
\usepackage{amsmath,amssymb,amsfonts}
\usepackage{algorithmic}
\usepackage{graphicx}
\usepackage{textcomp}
\usepackage{xcolor}
\usepackage{eso-pic}

\usepackage{booktabs}
\usepackage{multirow}
\usepackage[utf8]{inputenc}    
\usepackage[T1]{fontenc}       
\usepackage{tipa}              
\usepackage{textcomp}          
\usepackage{algorithmic}
\usepackage{textcomp}
\newcommand{\bng}[1]{\textsf{#1}}  

\def\bng{\bngx}

\font\bngx=bang10

\def\*#1*#2{o\null{#2}{#1}}

\def\sh#1{\setbox0=\hbox{#1}%
     \kern-.02em\copy0\kern-\wd0
     \kern.04em\copy0\kern-\wd0
     \kern-.02em\raise.0433em\box0 }

\def\BibTeX{{\rm B\kern-.05em{\sc i\kern-.025em b}\kern-.08em
    T\kern-.1667em\lower.7ex\hbox{E}\kern-.125emX}}

\makeatletter
\def\ps@IEEEtitlepagestyle{%
    \def\@oddfoot{\mycopyrightnotice}%
    \def\@evenfoot{}%
}
\def\mycopyrightnotice{
  
}
\let\old@ps@IEEEtitlepagestyle\ps@IEEEtitlepagestyle
\def\confheader#1{%
    \def\ps@IEEEtitlepagestyle{%
        \old@ps@IEEEtitlepagestyle%
        \def\@oddhead{\strut\hfill#1\hfill\strut}%
        \def\@evenhead{\strut\hfill#1\hfill\strut}%
    }%
    \ps@headings%
}
\makeatother

\makeatletter
\newcommand{\linebreakand}{%
  \end{@IEEEauthorhalign}
  \hfill\mbox{}\par
  \mbox{}\hfill\begin{@IEEEauthorhalign}
}
\makeatother

\begin{document}
\newcommand\AtPageUpperMyright[1]{\AtPageUpperLeft{
 \put(\LenToUnit{0.28\paperwidth},\LenToUnit{-1cm}){
     \parbox{0.78\textwidth}{\raggedleft\fontsize{9}{11}\selectfont #1}}
 }}

\title{Adaptability of ASR Models on Low-Resource Language: A Comparative Study of Whisper and Wav2Vec-BERT on Bangla\\


}

\author{
    \IEEEauthorblockN{Md Sazzadul Islam Ridoy,
                      Sumi Akter and
                      Md. Aminur Rahman} \\
    \IEEEauthorblockA{Department of Computer Science and Engineering \\
                      Ahsanullah University of Science and Technology, Dhaka, Bangladesh \\
                      Email: \{isazzadul23, sumi72541, aminur.rahman.rsd\}@gmail.com}
}




\maketitle

\begin{abstract}
In recent years, neural models trained on large multilingual text and speech datasets have shown great potential for supporting low-resource languages. This study investigates the performances of two state-of-the-art Automatic Speech Recognition (ASR) models, OpenAI's Whisper (Small \& Large-V2) and Facebook's Wav2Vec-BERT on Bangla, a low-resource language. We have conducted experiments using two publicly available datasets: Mozilla Common Voice-17 and OpenSLR to evaluate model performances. Through systematic fine-tuning and hyperparameter optimization, including learning rate, epochs, and model checkpoint selection, we have compared the models based on Word Error Rate (WER), Character Error Rate (CER), Training Time, and Computational Efficiency. The Wav2Vec-BERT model outperformed Whisper across all key evaluation metrics, demonstrated superior performance while requiring fewer computational resources, and offered valuable insights to develop robust speech recognition systems in low-resource linguistic settings.

\end{abstract}


\section{Introduction}
In recent years, sequence-based\cite{prabhavalkar2017comparison} models have revolutionized in speech recognition by using neural networks to map speech directly to text, significantly simplifying the process. Among sequence-based models, Transformer \cite{wolf2020huggingfacestransformersstateoftheartnatural,jaiswal2020survey} has shown remarkable success in building end-to-end speech recognition systems. Due to the lack of high-quality annotated speech data, Automatic Speech Recognition (ASR) for Bangla is considered as a low-resource language, making it difficult to train accurate models. Furthermore, the language has a complex orthographic system with diacritics, conjunct characters and regional phonetic variations\cite{paul2009bangla} in regional dialects, making speech-to-text mapping more challenging. That means a data-efficient method is imperative for the development of robust Bangla ASR\cite{kibria2022bangladeshi} systems. Recent breakthroughs in self-supervised learning\cite{jaiswal2020survey} have shown great promise in tackling the problem of limited data for underrepresented languages. Models like Wav2Vec-BERT \cite{chung2021w2vbertcombiningcontrastivelearning} leverage large amounts of unlabeled speech data to learn audio patterns, requiring only minimal labeled data for fine-tuning. This method reduces the need for massive annotated datasets while still boosting ASR accuracy by learning from raw data. However, to get the best results, self-supervised models need careful fine-tuning.
 On the other hand, fully supervised models\cite{graham2024evaluating} like OpenAI's Whisper are trained on huge multilingual datasets, allowing them to work well in different languages without much fine-tuning. Although Whisper is known for its impressive zero-shot capabilities \cite{graham2024evaluating}, its performance in low-resource languages like Bangla has not been explored in depth yet. This study makes several significant contributions by comparing Whisper variants (small \& large-v2) and Wav2Vec-BERT for Bangla ASR, focusing on accuracy (evaluated using metrics such as WER, CER, Training Time, and Computational Cost). We have examined model scalability by testing with two publicly available datasets: Mozilla Common Voice 17 \cite{commonvoice:2020} and OpenSLR\cite{kjartansson-etal-sltu2018, kjartansson-etal-tts-sltu2018} using five different dataset sizes ranging from 2,000 to 70,000 samples to evaluate how well each model handles different amounts of training data. We have run the models on two different computers with varying GPU, CPU, and RAM capacity to evaluate how these hardware differences affect training performance.

To the best of our knowledge, this is the first comprehensive analysis to directly evaluate Whisper and Wav2Vec-BERT on Bangla speech recognition, shedding light on their strengths and limitations in low-resource settings. This research not only advances the understanding of ASR models for Bangla but also bridges the gap in speech technology accessibility for Bangla speakers, enabling broader applications in education, healthcare, accessibility and governance.

\section{Background and Related Works}
\vspace{1.25mm}
\subsection{Bangla ASR}
Bangla is an Indo-Aryan language that consists of 11 vowels ({\bng sWorbor/No}) and 39 consonants ({\bng  bYNJ/jnbr/N}). The script is encoded in UTF-8 and follows an Abugida writing system \cite{paul2009bangla}, where each consonant has an inherent vowel sound ("{\bng A}" ô, known as {\bng s/br A} shôrô ô, pronounced /ô/), which can be modified using diacritics. Bangla also features complex consonant clusters, known as {\bng Juk/takKr } (Juktakkhor), significantly impacting pronunciation and speech recognition. For example, the cluster {\bng  j/jW } is formed by combining {\bng j} (\v{j}\^{o}), {\bng j} (\v{j}\^{o}) and {\bng b } (bô), resulting in the pronunciation (\v{j}\v{j} bô). Additionally, non-alphabetic characters such as {\bng AnusWor } (Anusshar, "{ \bng NNG }"), {\bng ibsor/g } (Visarga, "{ \bng h }") and {\bng cn/dRo ibn/du } (Chandrabindu, " { \bng NN }") play crucial roles in Bangla phonetics and orthography \cite{kibria2022bangladeshi}. These elements introduce nasalization and aspiration, further increasing the complexity of Bangla ASR systems. Recent efforts in Bangla speech recognition include the development of multiple speech corpora, such as the Bengali Common Voice dataset and the OpenSLR Bengali corpus. These resources have been instrumental in training and evaluating automatic speech recognition (ASR) systems.

\subsection{Wav2Vec-BERT}
Wav2Vec-BERT is an advanced speech recognition model that builds on Wav2Vec 2.0’s \cite{baevski2020wav2vec20frameworkselfsupervised} self-supervised learning approach while adding BERT’s ability to understand the context from both directions. It uses a combination of Convolutional Neural Networks (CNNs)\cite{alzubaidi2021review} and Transformers to process audio signals and learn meaningful linguistic patterns.
\begin{figure}[h!]
    \centering
    \includegraphics[width=0.5\textwidth]{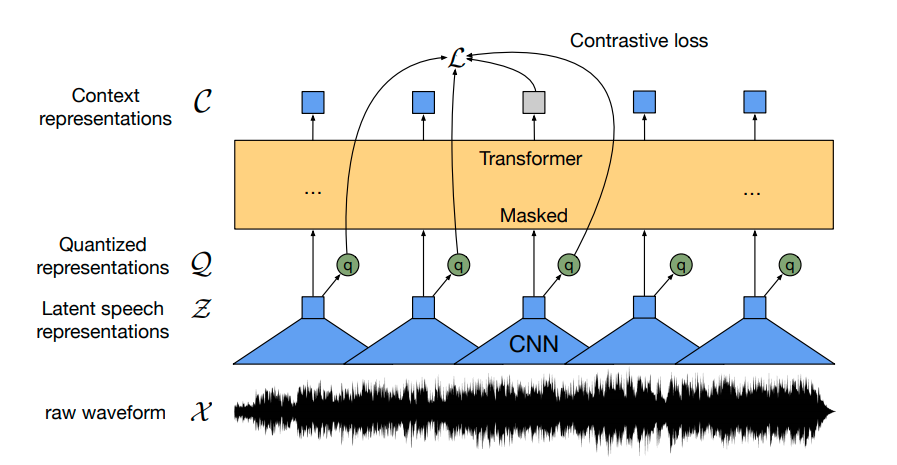}
    \caption{Wav2Vec 2.0 architecture representation \cite{baevski2020wav2vec20frameworkselfsupervised} }
    \label{fig:example-wv}
\end{figure}

Similar to Wav2Vec 2.0, as illustrated in Figure~\ref{fig:example-wv}, Wav2Vec-BERT begins by transforming raw audio \cite{jain2023wav2vec2} input $X$ into a latent speech representation $Z$ through a multi-layer convolutional encoder:
\[f: X \to Z\] These representations are then passed through a Transformer-based masked prediction network, which generates contextual embeddings, $c_1, \ldots, c_T$, by learning to predict masked portions of speech data: \[g: Z \to C\] Unlike its predecessor, Wav2Vec-BERT uses a bidirectional Transformer\cite{jaiswal2020survey} similar to BERT, allowing it to capture dependencies across the entire sequence instead of just left-to-right context. The model architecture includes a Conformer-based adapter\cite{li2023modular} network instead of a 
simple convolutional network. These representations are then discretized and passed into a BERT-style Transformer, which is pre-trained using a masked speech prediction objective. This helps the model to learn robust audio representations by reconstructing masked speech segments, improving its generalization across different languages and speech conditions. Wav2Vec 2.0 has shown better performance\cite{jain2023wav2vec2} than previous self-supervised ASR models, setting new records on several benchmark datasets. By combining the strengths of Wav2Vec 2.0 and BERT\cite{devlin2019bertpretrainingdeepbidirectional}, it significantly improves speech recognition, especially for low-resource languages where labeled data is limited.

\subsection{Whisper for ASR}
\begin{figure}[h!]
    \centering
    \includegraphics[width=0.5\textwidth]{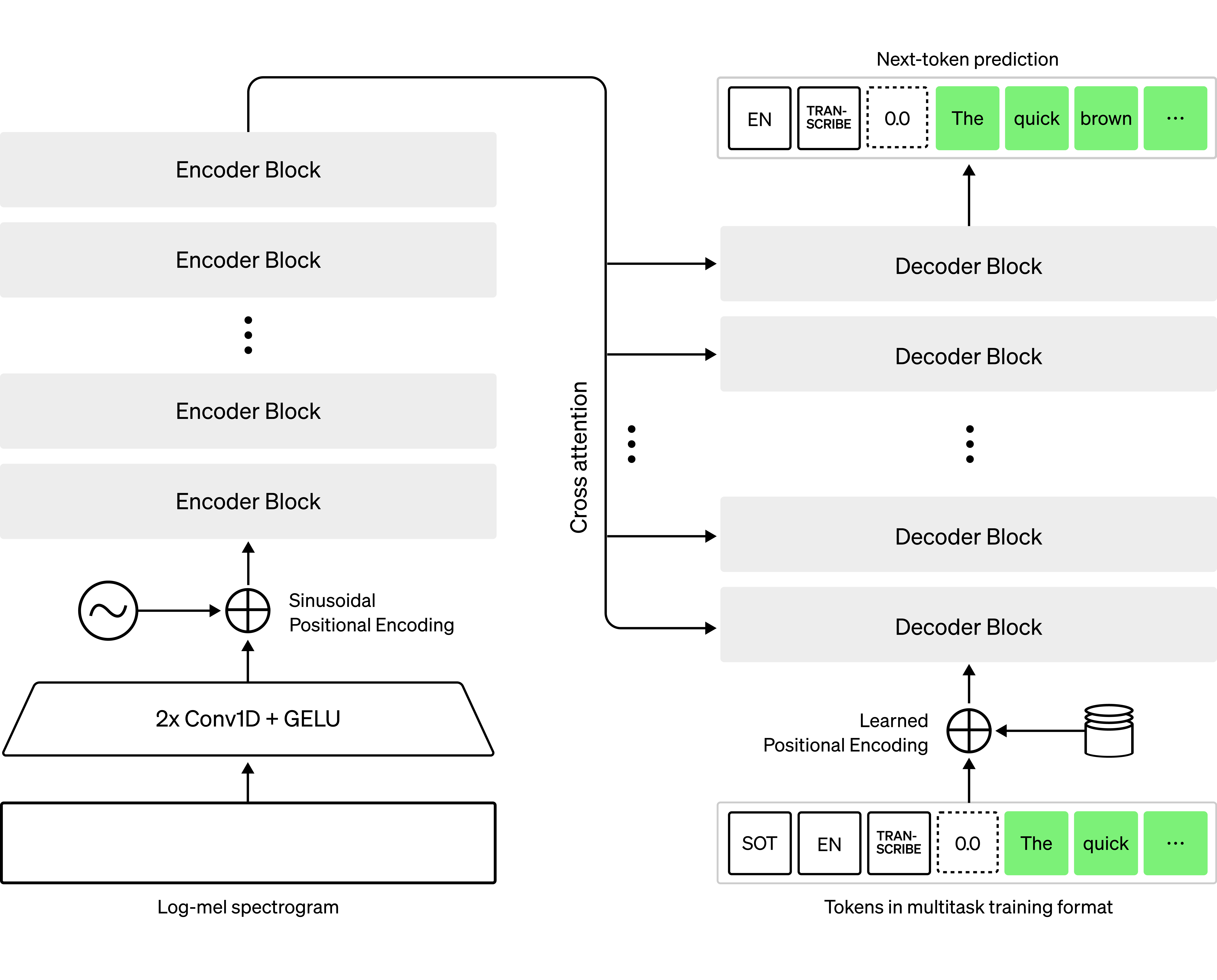}
    \caption{ASR Summary Of Whisper Model Architecture \cite{radford2022whisper} }
    \label{fig:example-w}
\end{figure}

The Whisper model, developed by OpenAI, represents another milestone in ASR research. Trained on an unprecedented 680,000 hours of labeled speech data, Whisper leverages a Transformer-based encoder-decoder architecture to handle multilingual and multitask speech processing. The model utilizes 80-channel log-Mel spectrograms\cite{kozhirbayev2023kazakh} as input, with the encoder consisting of two convolutional layers, sinusoidal positional encoding and a series of stacked Transformer blocks\cite{alzubaidi2021review}. In Figure \ref{fig:example-w}, the decoder employs learned positional embeddings and mirrors the encoder's architecture.
 Unlike Wav2Vec-BERT\cite{chung2021w2vbertcombiningcontrastivelearning}, Whisper\cite{radford2022whisper} adopts a fully supervised training approach, which involves using large amounts of annotated data. However, its reliance on labeled data makes it less adaptable to low-resource languages without extensive annotation efforts.
\vspace{1.25mm}
\section{Methodology}
\vspace{1.25mm}
Transformer architectures have become the leading approach for Automatic Speech Recognition (ASR) systems, especially when dealing with languages that have limited resources. These models can be implemented in two ways: training from scratch on massive amounts of annotated and unannotated data or fine-tuning pre-trained models like Wav2Vec-BERT and Whisper on annotated datasets\cite{singh2023model}. Due to the limited availability of unannotated Bangla speech data, this study focuses on fine-tuning Wav2Vec-BERT and Whisper (Small and Large v2) on publicly available Bangla speech datasets. This section outlines the datasets used, data processing techniques and the fine-tuning methodology.

\subsection{Datasets}
\subsubsection*{Annotated Speech Corpora}
Supervised deep learning techniques\cite{Gong_2023} require audio files paired with corresponding transcriptions to minimize the loss function and optimize model weights using backpropagation. Both Wav2Vec-BERT and Whisper rely on high-quality annotated datasets for effective fine-tuning.
\begin{itemize}
    \item Mozilla Common Voice (Bangla Subset):
    Version 17 of this dataset was released in March 2024 which includes 54 hours of verified annotated speech from 22,913 speakers and approximately 8 hours of unvalidated recordings. The dataset comprises 24,730 unique prompts. Given the total duration, this result is a relatively high repetition rate compared to other speech corpora that prioritize greater textual diversity. While the repetition limits linguistic variety, it helps maintain consistency in pronunciation across different speakers.
    \item OpenSLR Bangla Speech Dataset:
    This dataset offers approximately 40 hours of annotated speech with 27,308 unique prompts, covering diverse accents and recording conditions. It is widely used in academic research due to its high quality and comprehensive coverage.
\end{itemize}
The total annotated Bangla speech data used in this study amounts to approximately 86 hours, divided into training, validation and test sets. This diverse collection ensures a robust evaluation of the models across different domains and speaking styles.

\subsection{Data Processing}
To ensure consistency and enhance recognition accuracy, all audio files were resampled from 16 kHz to 8 kHz and then back to 16 kHz mono WAV format as a form of augmentation \cite{jain2023wav2vec2}. This process intentionally introduces the loss of high-frequency components and potential quantization noise, which can help improve model robustness. Preprocessing on both the text and audio sides included:

\begin{itemize}
    \item Text Normalization: Expanding abbreviations, removing unnecessary punctuations (except apostrophes) and converting numbers into Bangla words (e.g., {\bng "123"} → {\bng Eksh etIsh }) for consistency in spoken form.
    \item Audio Pre-Processing:
    For Wav2Vec-BERT, raw waveforms were normalized within a [-1, 1] amplitude range to enhance robustness across different recording environments. Subword tokenization \cite{radford2022whisper} was performed using a default vocabulary mapping function. For Whisper, audio was converted into log-Mel spectrograms with a 30 ms window and a 10 ms stride to match its expected input format. Since Whisper supports multilingual ASR\cite{radford2022whisper}, the text was labeled with the appropriate language token (e.g., <|bn|> for Bangla). Unlike Wav2Vec-BERT, Whisper does not require explicit forced alignment due to its end-to-end training on large-scale paired text-audio data.
        
\end{itemize}
\begin{figure}[h!]
    \centering
    \includegraphics[width=0.5\textwidth]{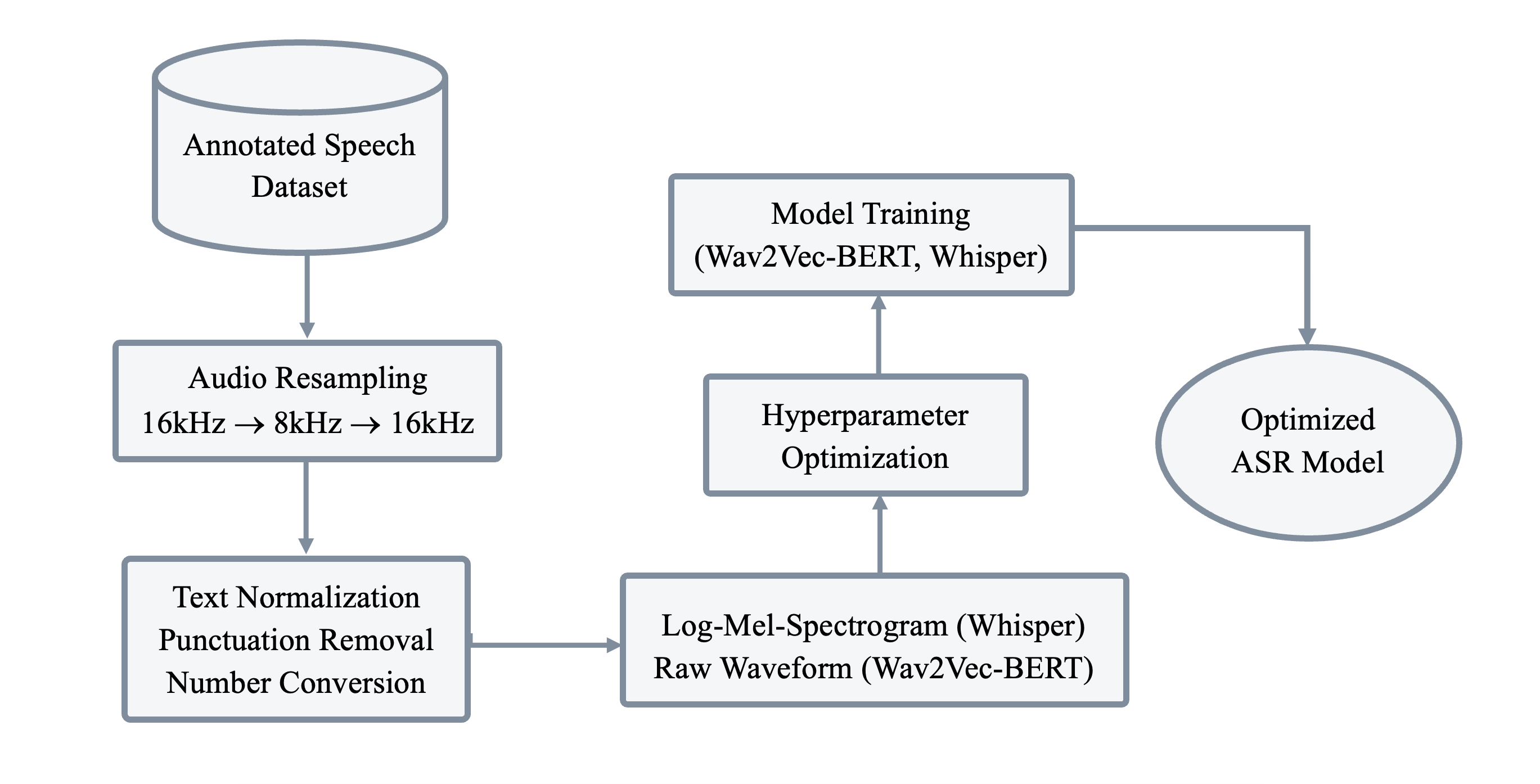}
    \caption{Architecture of the Fine-Tuning Process for Bangla ASR using Wav2Vec-BERT and Whisper}
    \label{fig:example}
\end{figure}

\subsection{Dataset Subsets and Scaling Analysis}
To thoroughly evaluate the impact of dataset size on model performance, the annotated Bangla speech data was divided into five subsets: 2k, 8k, 20k, 40k and 70k samples. Both models were then fine-tuned separately using these five dataset sizes to assess their performance. This approach enabled a comprehensive analysis of how the models respond to varying amounts of training data and revealed incremental performance improvements. To enhance accuracy, especially in smaller subsets, we prioritized unique prompts by filtering the data to eliminate duplicates. This strategy ensured a diverse representation of speech patterns and minimized overfitting. The uniqueness of prompts was calculated using the following formula:
\[
|U| = \left| \{ t \in T \mid t \text{ appears at least once in } T \} \right|
\]

Where:  
\begin{itemize}
    \item \( T = \{ t_1, t_2, \ldots, t_n \} \) is the set of all transcriptions (prompts) in the dataset, where \( t_i \) represents the transcription of the \( i^{\text{th}} \) utterance.
    \item \( U \) is the set of unique transcriptions.
    \item \( |U| \) denotes the cardinality (size) of the set \( U \), indicating the total number of unique prompts.
\end{itemize}
This methodical approach to dataset scaling and unique prompt selection provided a robust foundation for understanding how data diversity influences model learning and accuracy in Bangla ASR systems.

\subsection{Model Fine-Tuning}
We fine-tuned the Wav2Vec-BERT and Whisper models, including both the Small and Large-v2 variants of Whisper for Bangla ASR, using the Mozilla Common Voice and OpenSLR Bangla speech datasets. Five different dataset sizes were used during the fine-tuning process to evaluate the impact of data scaling on model performance.

\subsubsection{Hardware Configurations}
Fine-tuning was performed on two hardware setups to understand the effect of computational resources on training time and model performance:
\begin{itemize}
    \item High-End Setup: NVIDIA RTX 4090 GPU (24 GB VRAM) – Allowed faster training with larger batch sizes.
    \item Low-End Setup: NVIDIA RTX 3060 GPU (12 GB VRAM) – Provided a resource-constrained environment to generalize memory usage, training time and inference speeds across different hardware configurations.
\end{itemize}

\subsubsection{Hyperparameter Tuning}
Three sets of hyperparameters were tested for each model, varying epoch size, learning rate, step size and evaluation frequency. The configurations were as follows:
\begin{itemize}
    \item Epochs: For smaller datasets (2k and 8k samples), 10 and 15 epochs were used, while for larger datasets (20k, 40k, and 70k samples), 8 and 10 epochs were employed to avoid overfitting.
    \item Learning Rate: Wav2Vec-BERT was fine-tuned with an initial learning rate of $3 \times 10^{-5}$, while Whisper used $1 \times 10^{-5}$. Both models employed a warm-up schedule for the first 500 steps.
    \item Batch Size and Gradient Accumulation: Batch sizes were adjusted based on model size and GPU memory. Wav2Vec-BERT utilized larger batch sizes, whereas Whisper required smaller batches. Gradient accumulation was performed every four steps for efficient memory utilization in larger models such as Whisper Large-v2.
\end{itemize}
This fine-tuning approach, across multiple datasets, hardware setups and hyperparameter settings, provides a comprehensive evaluation of Wav2Vec-BERT and Whisper for Bangla ASR. It also exposes the strengths and weaknesses of these models, particularly in low-resource language scenarios.

\section{Result and Discussion}
\vspace{1.25mm}
The performances of Wav2Vec-BERT and Whisper models were evaluated using different dataset sizes and computational setups. The experiments revealed significant variations in Word Error Rate (WER) and Character Error Rate (CER), influenced by the model architecture, dataset scale and hardware configuration.

\begin{figure}[h!]
    \centering
    \includegraphics[width=0.5\textwidth]{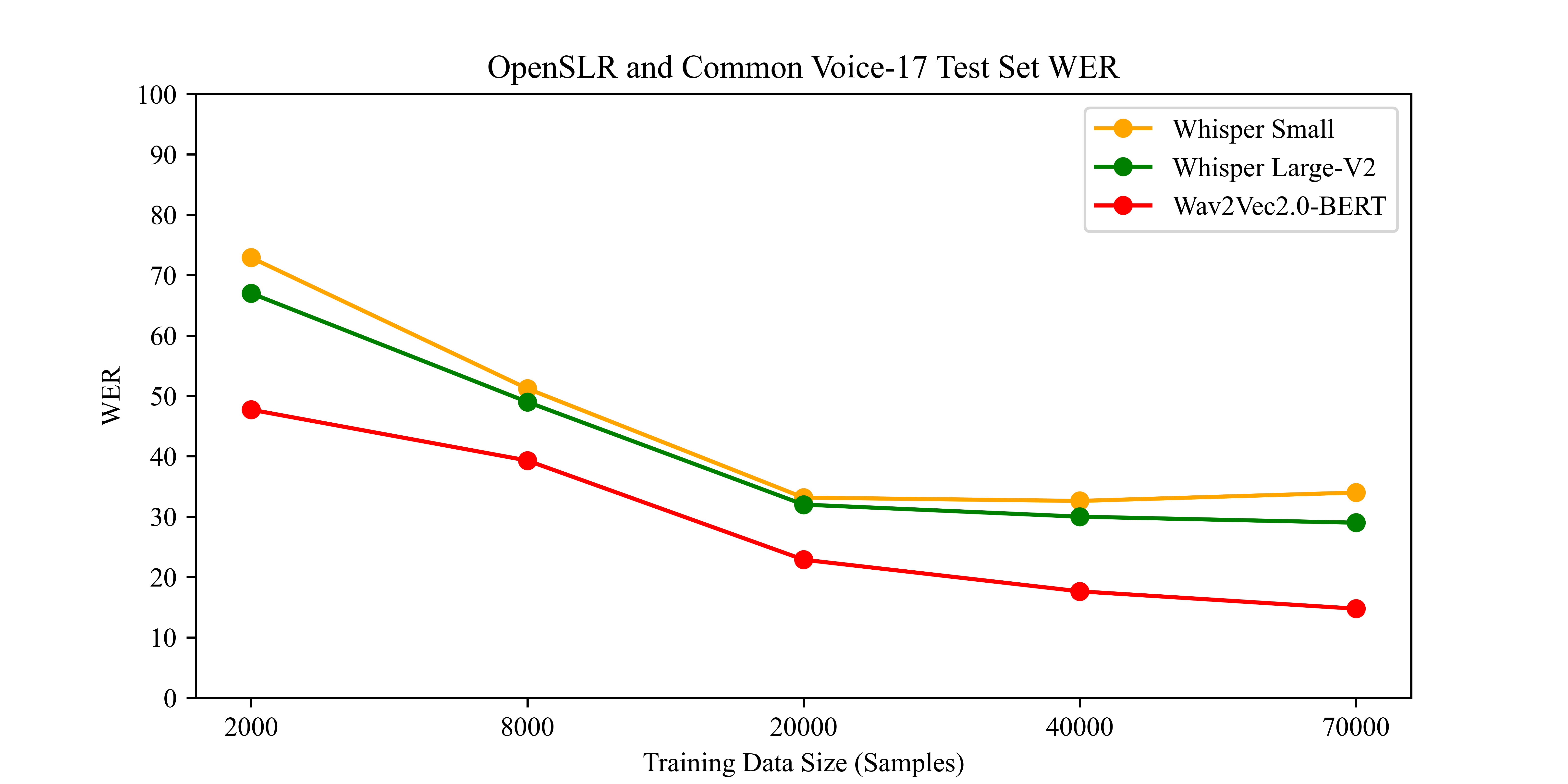}
    \caption{Wav2Vec-BERT and Whisper WER Results on the OpenSLR and Common Voice -17 Test Set}
    \label{fig:wer}
\end{figure}

\begin{table*}[h]
\centering
\caption{Impact of Learning Rate and Epochs on WER (Dataset Size: 70k, GPU: RTX 4090)}
\label{tab:my-table}
\begin{tabular}{|c|c|c|c|c|c|}
\hline
\textbf{MODEL}                    & \textbf{LEARNING RATE} & \textbf{EPOCHS} & \textbf{WER (\%)} & \textbf{CER (\%)} & \textbf{TRAINING TIME (HH:MM)} \\ \hline
\multirow{3}{*}{Wav2Vec-BERT}     & 1e-5                   & 8               & \textbf{14.42}    & \textbf{2.67}     & \textbf{13:26}                 \\ \cline{2-6} 
                                  & 3e-5                   & 10              & 17.61             & 3.04              & 14:04                          \\ \cline{2-6} 
                                  & 5e-5                   & 15              & 72.31             & 21.93             & 17:37                          \\ \hline
\multirow{3}{*}{Whisper-small}    & 1e-5                   & 8               & 32.73             & 19.58             & 15:39                          \\ \cline{2-6} 
                                  & 3e-5                   & 10              & 33.91             & 18.93             & 17:53                          \\ \cline{2-6} 
                                  & 5e-5                   & 15              & 32.28             & 18.31             & 21:47                          \\ \hline
\multirow{3}{*}{Whisper-large-v2} & 1e-5                   & 8               & 29.43             & 9.26              & 19:12                          \\ \cline{2-6} 
                                  & 3e-5                   & 10              & 28.86             & 7.47              & 21:52                          \\ \cline{2-6} 
                                  & 5e-5                   & 15              & 31.36             & 8.81              & 25:21                          \\ \hline
\end{tabular}
\end{table*}

\begin{table*}[]
\centering
\caption{Best Model Configurations}
\label{tab:my-table-2}
\begin{tabular}{|c|c|c|c|c|c|c|}
\hline
\textbf{MODEL}   & \textbf{DATASET SIZE} & \textbf{EPOCHS} & \textbf{LEARNING RATE} & \textbf{WER (\%)} & \textbf{CER(\%)} & \textbf{\begin{tabular}[c]{@{}c@{}}TRAINING TIME \\ (HH:MM)\end{tabular}} \\ \hline
Wav2Vec-BERT & 70k samples           & 8              & 1e-5                   & \textbf{14.42}    & \textbf{2.67}     & 13:26                                                                     \\ \hline
Whisper Small    & 40k samples           & 15             & 1e-5                   & 32.17             & 18.17              & 16:13                                                                     \\ \hline
Whisper Large-v2 & 70k samples           & 10             & 3e-5                   & 28.86             & 7.47              & 21:52                                                                     \\ \hline
\end{tabular}
\end{table*}

\begin{table*}[h]
\centering
\caption{Common errors in Whisper and Wav2Vec-BERT}
\label{tab:errors}
\begin{tabular}{|c|c|c|c|}
\hline
\textbf{True Text} & \textbf{Whisper} & \textbf{Wav2Vec-BERT} & \textbf{Common Error Type}                                       \\ \hline
\bng ibShN/N       & \bng ibShN/N     & \bng ibShn/n     & Context-sensitive position confusion (Wav2Vec-BERT)  \\ \hline
\bng jhrh          & \bng Jrh         & \bng jhrh        & Voiced retroflex confusion (Whisper)                             \\ \hline
\bng shsYo         & \bng ssYo        & \bng shsYo       & Voiceless fricative confusion (Whisper)                          \\ \hline
\bng tebla         & \bng thebla      & \bng tebla       & Aspirated/unaspirated mismatch (Whisper)                         \\ \hline
\bng 8 I           & \bng AaiT        & \bng AaT I       & Numeral-word misinterpretation (Whisper)                         \\ \hline
\end{tabular}
\end{table*}

\subsection{Wav2Vec-BERT Performance}
\begin{figure}[h!]
    \centering
    \includegraphics[width=0.5\textwidth]{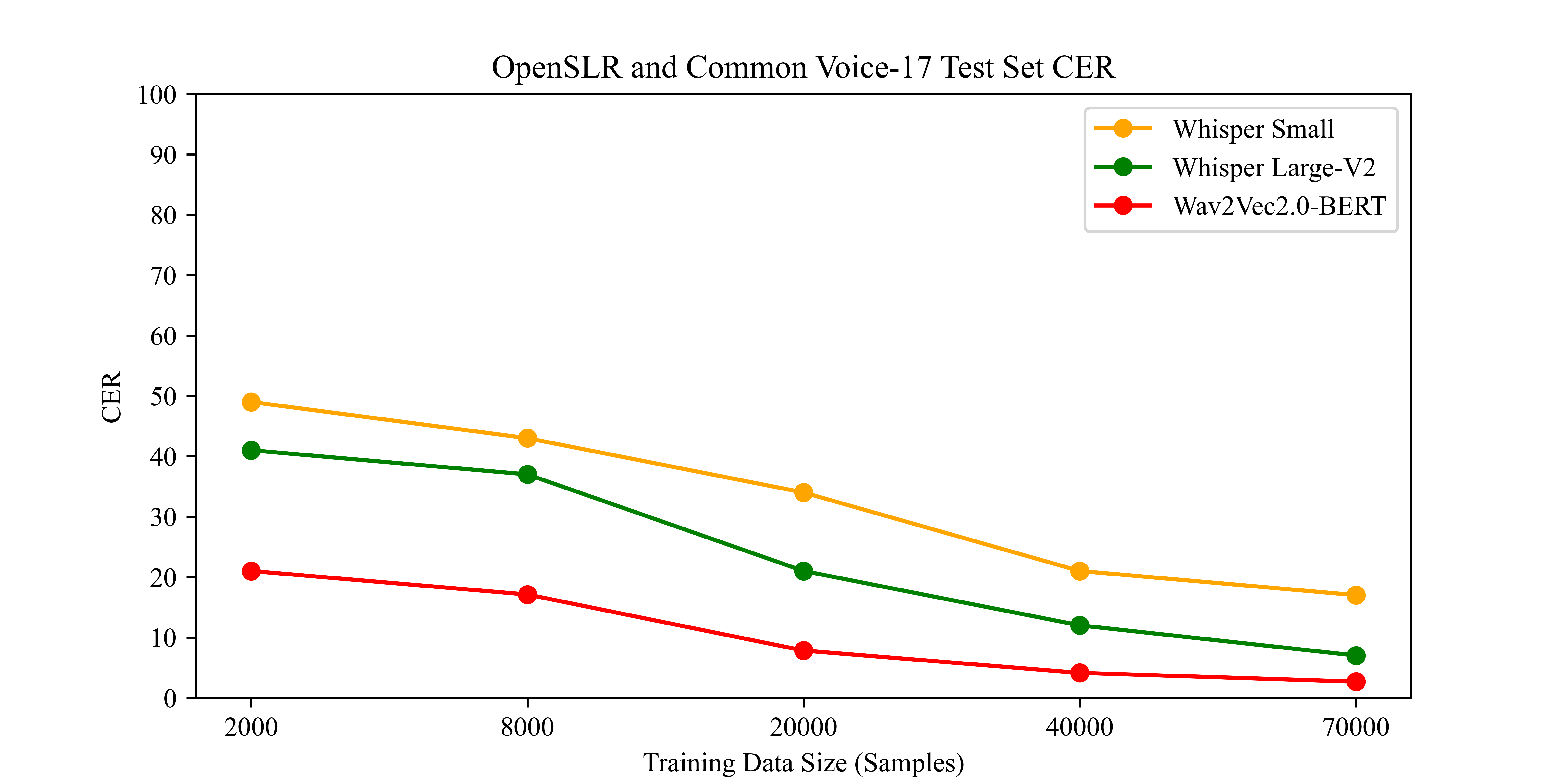}
    \caption{Wav2Vec-BERT and Whisper CER Results on the OpenSLR and Common Voice -17 Test Set}
    \label{fig:cer}
\end{figure}

The performance of Wav2Vec-BERT across various dataset sizes is illustrated in Figure~\ref{fig:wer} (WER trend) and Figure~\ref{fig:cer} (CER trend). Table~\ref{tab:my-table} shows the impact of different learning rates and epochs on model accuracy, while Table~\ref{tab:my-table-2} presents the best configuration per model. In Figure~\ref{fig:wer}, the WER curve flattens after 40k samples, indicating diminishing returns on further data increase. This aligns with Table~\ref{tab:my-table-2}, where 70k samples and 8 epochs yield optimal performance (WER 14.42\%, CER 2.67\%). Table~\ref{tab:my-table} reveals a clear overfitting pattern at 15 epochs (WER jumps to 72.31\%), demonstrating sensitivity to training duration. These trends collectively underscore the importance of hyperparameter tuning and controlled training for Wav2Vec-BERT in low-resource settings. A key advantage of Wav2Vec-BERT is its eﬀicient utilization of computing resources. It successfully completed training on both lower-end and high-end PCs without encountering memory constraints, highlighting its versatility and lower VRAM requirements.

\subsection{Whisper Model Analysis}
Whisper models, despite their advanced architecture, showed higher resource demands. Whisper Small achieved the best WER of 32.61\% and CER of 18.17\% on the 40k dataset but encountered memory issues on lower-end PCs when processing datasets larger than 20k. Using the 70k dataset, Whisper Large-v2 achieved a WER of 28.86\% alongside a CER of 7.47\%, but it required a high-end PC due to its substantial VRAM and RAM consumption. Whisper model's extensive VRAM requirements are mainly due to its detailed attention mechanisms and dense layers designed for precise audio mapping.

\subsection{Statistical Significance Testing}
To ensure observed differences in model performances
were not due to random variation, we conducted paired
$t$-tests on WER and CER values across identical dataset sizes and epochs for both models. A p-value measures the likelihood that the results occurred under the null hypothesis; values below 0.05 indicate statistically significant differences.

At 70k dataset size (8 epochs):

\begin{itemize}
    \item WER comparison (Whisper Large-v2 vs. Wav2Vec-BERT): $p = 0.0041$
    \item CER comparison: $p = 0.0037$
\end{itemize}
These results confirm that Wav2Vec-BERT significantly outperforms Whisper in both WER and CER metrics ($p < 0.05$).

\subsection{Error Analysis and Common Mistakes}

To provide qualitative insights, we analyzed phoneme and grapheme-level errors using a confusion matrix derived from 500 test utterances, summarized in Table \ref{tab:errors}.

\subsection*{Wav2Vec-BERT Confusions}
\begin{itemize}
    \item {\bng n }(dental nasal \textipa{n}) vs. {\bng N} (retroflex nasal \textipa{N}): Errors occurred primarily in context-sensitive positions like compound words or loanwords.
\end{itemize}

\subsection*{Whisper Confusions}
\begin{itemize}
    \item {\bng jh} (jho) vs. {\bng J} (\textipa{j}): The model often failed to differentiate due to similar voicing and articulation patterns.
    
    \item {\bng sh} (\textipa{S}) vs. {\bng  s} (\textipa{s}): Common substitutions likely due to acoustic similarity in fricative production.
    
    \item {\bng t } (\textipa{t}) vs. {\bng th } (\textipa{t\textsuperscript{h}}): The aspirated/unaspirated distinction was inconsistently recognized, particularly in fast speech.
    
\end{itemize}

\subsection*{Common Word-Level Errors:}
\begin{itemize}
\item Whisper frequently exhibited confusion with voiced and voiceless consonants (e.g., retroflex and fricative sounds), misinterpreted aspirated vs. unaspirated phonemes, and struggled with numeral-word combinations.
\item Wav2Vec-BERT showed context-sensitive positional confusion, particularly in phoneme boundary recognition (e.g., nasal endings), but was more accurate with fricatives and numerals than Whisper.
\end{itemize}

\subsection*{Comparative Evaluation:}
While both models exhibit confusions between phonetically similar sounds, Wav2Vec-BERT shows errors in nasal consonant distinctions, whereas Whisper struggles more with fricative and aspirated/unaspirated pairs. Whisper also sometimes makes errors when converting numbers into Bangla words. This comparison shows that each model has different types of weaknesses, which could help guide future improvements focused on these specific sound challenges.

\section{Conclusion}
\vspace{1.25mm}
This study presents a comparative analysis of two Bangla automatic speech recognition (ASR) models, Wav2Vec-BERT and Whisper, highlighting their respective strengths, limitations, and error patterns. Experiments were conducted with different dataset sizes, training times and hardware setups, helping to understand how these models perform for Bangla. Wav2Vec-BERT proved highly adaptable, making it a great option for resource-constrained environments. It was efficient in training and resource usage, performing well across different hardware setups. Whisper models kept improving with more data. However, they required more computing power and memory and the large-v2 model failed to run on a low-end setup. This comparison helps guide the choice between Wav2Vec-BERT and Whisper for Bangla ASR, balancing efficiency, accuracy and resource requirements. Wav2Vec-BERT demonstrates higher overall accuracy and efficiency, making it a more suitable option for general use, while Whisper requires more computational resources but does not consistently outperform Wav2Vec-BERT in terms of accuracy. Overall, this study provides practical guidance for building Bangla ASR systems. It also contributes to multilingual ASR research, showing how advanced models can work for low-resource languages. Future work should focus on improving real-world usability, refining training methods and expanding high-quality annotated datasets to continue advancing Bangla ASR technology.

\bibliographystyle{IEEEtran}
\bibliography{ref}

\end{document}